# DANCER: Entity Description Augmented Named Entity Corrector for Automatic Speech Recognition

Yi-Cheng Wang[1,*], Hsin-Wei Wang[1], Bi-Cheng Yan[1], Chi-Han Lin[2] and Berlin Chen[1,*]
[1]National Taiwan Normal University, Taipei, Taiwan
[2]E.SUN Financial Holding Co., Ltd., Taipei, Taiwan
[1]{yichengwang, hsinweiwang, bicheng, berlin}@ntnu.edu.tw
[2]finalspaceman-19590@esunbank.com

## Abstract

End-to-end automatic speech recognition (E2E ASR) systems often suffer from mistranscription of domain-specific phrases, such as named entities, sometimes leading to catastrophic failures in downstream tasks. A family of fast and lightweight named entity correction (NEC) models for ASR have recently been proposed, which normally build on phonetic-level edit distance algorithms and have shown impressive NEC performance. However, as the named entity (NE) list grows, the problems of phonetic confusion in the NE list are exacerbated; for example, homophone ambiguities increase substantially. In view of this, we proposed a novel **D**escription **A**ugmented **N**amed entity **C**or-r**E**cto**R** (dubbed DANCER), which leverages entity descriptions to provide additional information to facilitate mitigation of phonetic confusion for NEC on ASR transcription. To this end, an efficient entity description augmented masked language model (EDA-MLM) comprised of a dense retrieval model is introduced, enabling MLM to adapt swiftly to domain-specific entities for the NEC task. A series of experiments conducted on the AISHELL-1 and Homophone datasets confirm the effectiveness of our modeling approach. DANCER outperforms a strong baseline, the phonetic edit-distance-based NEC model (PED-NEC), by a character error rate (CER) reduction of about 7% relatively on AISHELL-1 for named entities. More notably, when tested on Homophone that contains named entities of high phonetic confusion, DANCER offers a more pronounced CER reduction of 46% relatively over PED-NEC for named entities. The code is available at https://github.com/Amiannn/Dancer.

## 1. Introduction

Owing to the ease of scalability and the monolithic nature of end-to-end (E2E) neural models, general-purpose E2E automatic speech recognition (ASR) models have seen remarkable success and surpassed conventional hybrid ASR models across a multitude of domains (Radford et al., 2023). However, a well-known vulnerability of E2E ASR (Sainath et al., 2018) models is that they often cause mistranscription of domain-specific words or phrases, such as named entities (NEs), which occur infrequently in the training set. These NEs include personal names, locations, organizations, product names, and more, which are often transcribed into more common words in the vocabulary.

In recent years, a significant body of work has strived to address the NEC issue for ASR, which broadly falls into three categories: 1) post-training integration of external language models (LMs); 2) training-time integration of domain-specific contexts; 3) post-correction on ASR transcripts. Mainstream NEC approaches in the first category involve incorporating domain-specific information into the ASR model, for example on-the-fly rescoring (Hall et al., 2015; Kim et al., 2019) with a domain-specific *n*-gram or neural language models. Yet another similar method is to perform shallow fusion with a domain-biased weighted finite-state transducer (WFST) (Williams et al., 2018; Pundak et al., 2018; Mohri, 2018). In the second category, popular methods include those stemming from neural contextual biasing (Bruguier et al., 2019; Chen et al., 2019; Jain et al., 2020; Le et al., 2021; Chang et al., 2021; Sathyendra et al., 2022). These methods encode a list of domain-specific phrases or named entities into contextual embeddings interacting through cross-attention mechanisms and jointly optimized with the ASR model. The third category, known as ASR error correction, focuses exclusively on refining ASR transcripts and rectifying errors in the text space. Methods of this category typically frame NEC for ASR as a machine translation problem following an autoregressive sequence-to-sequence (Seq2Seq) framework (Mani et al., 2020; Park et al., 2021; Wang et al., 2020). Despite the existing methods mentioned above have shown potential to significantly improve the performance of various domain-specific ASR tasks to varying extents, each of them still has certain limitations. For example, methods of the first category normally require vast amounts of domain-specific training data, making them tend to be data-intensive. As for methods of the second category, when the customized list of domain-specific phrases including hundreds to thousands of entities curated from a huge catalog, the computational requirements of these contextual biasing methods become very demanding since they typically bias the ASR model towards the correct entities through a cross-attention mechanism which is relatively computationally intensive. Finally, methods of the third category usually suffer from slow inference speed because of their autoregressive nature and may produce excessive or hallucinatory refinements since the models are only constrained by the text transcripts generated by an ASR model, causing the NEC results to deviate from the true lexical information of an input utterance.

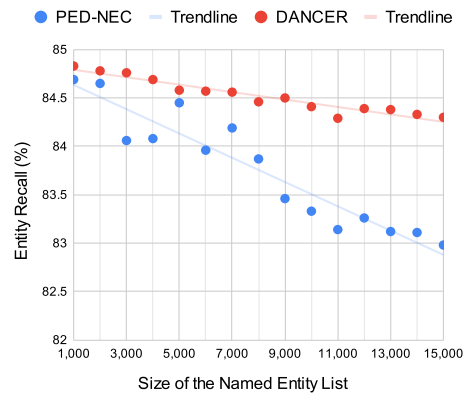

Figure 1: Impact of the named entity list size on AISHELL-1 test set. When scaling up the NE list, the problems of phonetic confusion in the NE list increase substantially. Our proposed DANCER model can effectively leverage entity semantics to alleviate this problem. There is a sizable performance gap between DANCER and the phonetic edit-distance-based NEC method (PED-NEC) as the entity sizes increase.

---

* Corresponding author.

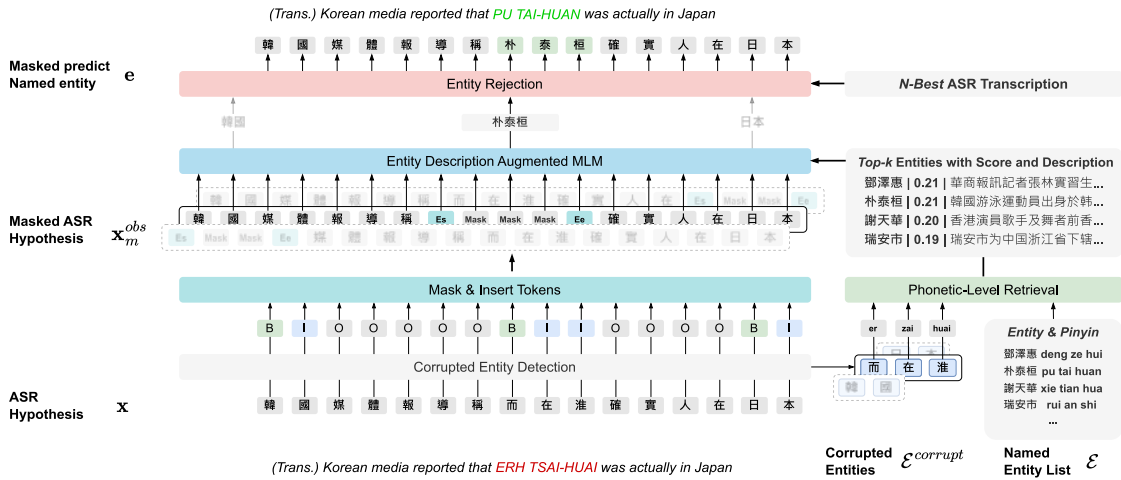

Figure 2: A schematic illustration of our proposed DANCER method.

Post-correction methods are considered the most generic and suitable approach to NEC, especially when using a production ASR system running on the cloud, which makes it intractable to alter the model components of the ASR system. Recent work on developing post-correction methods concentrates on compensating for the downsides of slow inference speed and the absence of acoustic constraints mentioned earlier by employing non-autoregressive models (Leng et al., 2021a; Leng et al., 2021b; Leng et al., 2023; Wang et al., 2022; Li, 2022), which involves forced-alignment between named entities and an input utterance (Kuo and Chen, 2022; Lin and Wang, 2023).
Among these, fast and lightweight methods based on textual and phonetic similarity computed by edit distance algorithms have shown remarkable NEC performance (Raghuvanshi et al., 2019; Garg et al., 2020). However, as the NE list is enlarged, the problems stemming from textual and phonetic confusion, especially for homophonic words or phrases, become substantially worse. We thus argue that incorporating the semantic meanings of named entities might help alleviate phonetic confusion. For this to work, the challenge boils down to how to effectively integrate entity knowledge into these models meanwhile balancing the contributions from semantic and phonetic matching information of named entities. Language models, particularly non-autoregressive masked language models (MLMs) (Devlin et al., 2019), can well encode the semantics of an entity; nonetheless, they store existing factual knowledge within their parameters solely, making it hard to adapt to unseen or rare named entities. In this paper, we present a novel description augmented named entity corrector, dubbed

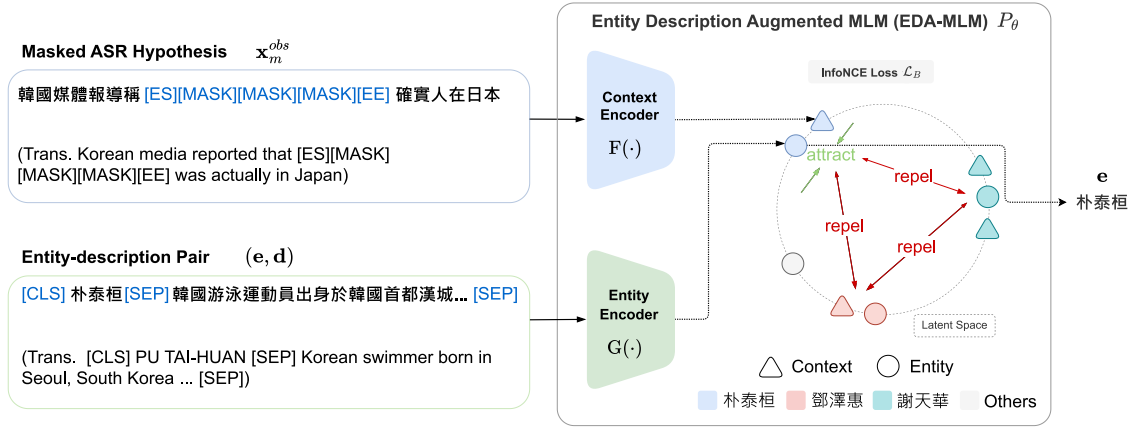

Figure 3: A schematic illustration of the proposed entity description augmented masked language model (EDA-MLM).

DANCER, which leverages entity descriptions[1] to tackle the critical issue that the adverse impact of phonetic confusion in the NE list will become even worse when the NE list increases, as shown in Figure 1. Notably, our work also inherits the merits of the fast and lightweight edit-distance-based NEC model. On this basis, we further develop an efficient entity description augmented masked language model (EDA-MLM) to work around the inherent phonetic confusion among named entities in the NE list. EDA-MLM is composed of a dense retrieval model and entity-description memories, empowering it to adapt and accommodate well to unseen entities. A series of experiments conducted on the AISHELL-1 (Bu et al., 2017) and Homophone benchmark datasets seem to validate the effectiveness and practicality of our method and its associated modeling approaches.

## 2. Methodology

In this work, we focus primarily on addressing the phonetic confusion problem facing the ASR NEC process by leveraging entity semantics and their fusion with other information cues. Figure 2 illustrates the overview architecture of our DANCER model. We begin by elaborating on the problem formulation and the objective of our DANCER model. In the subsequent subsections, we first discuss the strategy for detecting corrupted entities and then flesh out the model component for performing phonetic-level retrieval of entities. We then delve into the details of the entity description augmented MLM (EDA-MLM). After that, we introduce a hyperparameter to regulate the balance between the use of semantics and phonetic information of named entities for ASR NEC. Furthermore, we equip the DANCER model with an entity rejection mechanism to ensure it does not radically replace named entities that are already correctly recognized.

### 2.1 Overview

Given an ASR hypothesis with $L$ character tokens represented by $\mathbf{x} = (x_1, \cdots, x_L)$, and a list of pre-defined NE list consisting of $C$ entities denoted as $\mathcal{E} = \{\mathbf{e}_1, \cdots, \mathbf{e}_C\}$, where $\mathbf{e}_c$ is an entity composed from a list of character tokens. The goal of an NEC model is to identify the possible set of corrupted entities $\mathcal{E}^{corrupt} = (\mathbf{e}_1^{corrupt}, \cdots, \mathbf{e}_M^{corrupt})$ for the ASR hypothesis of a given utterance and replace each of them with their correct counterparts meticulously selected from the NE List $\mathcal{E}$. Before delving into the detail of our DANCER model, we first briefly review the phonetic edit-distance-based NEC (PED-NEC) model.

PED-NEC refines an ASR hypothesis $\mathbf{x}$ by first detecting all the corrupted entity $\mathbf{e}_m^{corrupt}$ in $\mathbf{x}$. PED-NEC then utilizes a phonetic-level matching mechanism based on the edit-distance algorithm to substitute the corrupted entity by matching the most phonetically similar entity picked from the NE list $\mathcal{E}$. The intuition behind PED-NEC is that ASR systems typically mistranscribe entities to acoustically similar words. Hence, a simple phonetic retrieval method is expected to be effective in achieving high performance. However, when the NE list increases, the phonetic confusion among entities in the NE list grows significantly, which leads to a significant performance drop of PED-NEC. To remedy this, our DANCER model takes the description of each entity into account, which provides additional semantic clues to alleviate the phonetic confusion. This involves providing a list of entity-description pairs $\mathcal{C} = \{(\mathbf{e}_c, \mathbf{d}_c)\}_{c=1...C}$ to our model (*cf.* section 3.2), where $\mathbf{d}_c$ is the corresponding description of an entity $\mathbf{e}_c$.

---

[1] We choose entity description since the text data is most easy to access, and the entity usually has a specific definition.

Our proposed DANCER proceeds in four stages: 1) a corrupted entity detection module is in charge of detecting the corrupted entities $\mathcal{E}^{corrupt}$ in the ASR hypothesis $\mathbf{x}$; 2) we sweep through all $\mathbf{e}_m^{corrupt}$ one by one and utilize a phonetic-level retriever to retrieve a subset of top-$k$ phonetically similar entity candidates $\mathcal{E}_\mathcal{F}$ from the NE list $\mathcal{E}$ with respect to $\mathbf{e}_m^{corrupt}$; 3) an EDA-MLM model is incorporated to rerank the entity candidates $\mathcal{E}_\mathcal{F}$ with their respective descriptions, the entity-description subset is denoted as $\mathcal{C}_\mathcal{F}$. It is also worth mentioning here that we introduce a hyperparameter $\alpha$ during the ranking process to regulate the contributions of the phonetic and semantic similarity scores, and we select the entity candidate with the highest holistic score to substitute the corrupted entity; and 4) an entity rejector is subsequently employed to avoid mistakenly replacing named entities that are already correctly recognized.

## 2.2 Corrupted Entity Detection

We formulate corrupted entity detection (CED) as a sequence classification task. Given an input of ASR hypothesis $\mathbf{x}$, the CED model, consisting of $U$ layers of Transformer (Vaswani et al., 2017) blocks, is in charge for categorizing each $x_i$ from $\mathbf{x}$ into three classes with the BIO format. In this format, B and I indicate the beginning and interior of the corrupted entity, while O represents an uncorrupted token. To optimize the CED model, we utilize cross-entropy loss.

## 2.3 Phonetic-level Retrieval

The phonetic-level retrieval component (i.e., the conditional probability $P_\eta$) can be approximated by applying $\text{Pin}(\cdot)^2$, and utilizing a normalized edit distance measure, $\text{ED}(\cdot)$, to calculate the phonetic similarity between $\mathbf{e}_c$ and $\mathbf{e}_m^{corrupt}$, as denoted by:

$$P_\eta(\mathbf{e}_c \mid \mathbf{e}_m^{corrupt}) = \frac{\text{ED}(\text{Pin}(\mathbf{e}_c), \text{Pin}(\mathbf{e}_m^{corrupt}))}{\sum_{\tilde{\mathbf{e}}_c \in \mathcal{E}} \text{ED}(\text{Pin}(\tilde{\mathbf{e}}_c), \text{Pin}(\mathbf{e}_m^{corrupt}))} . \quad (1)$$

## 2.4 Entity Description Augmented MLM

The non-autoregressive masked language model (MLM) is renowned for its efficiency and capacity to derive rich contextual representations for an anchor mask. However, one drawback that MLM faced with is its insufficient capacity of adapting and accommodating to unseen phrases, as MLM merely retains all known factual knowledge within its parameter space. Our workaround, entity description augmented MLM (EDA-MLM), draws some inspiration from (Khandelwal et al., 2020; Lewis et al., 2020; Fu et al., 2022; Jong et al., 2022; Wu et al., 2020) and employs a dense retrieval mechanism to aid MLM in retrieving semantically similar entity-description pairs for the masked corrupted entity whose embedding is derived based on its surrounding context.

The EDA-MLM model (i.e., the conditional probability $P_\theta$) is computed by a bi-encoder neural architecture comprising a context encoder $\text{F}(\cdot)$ and an entity encoder $\text{G}(\cdot)$. Figure 3 illustrates the architecture of EDA-MLM. The probability of $\mathbf{e}$ being the corresponding correct entity for the masked corrupted entity ASR hypothesis $\mathbf{x}_m^{obs}$ constrained on the entity-description candidates $\mathcal{C}_\mathcal{F}$ can be computed by:

$$\begin{aligned}&P_\theta(\mathbf{e} \mid \mathbf{x}_m^{obs}) \\ &= \frac{\exp(\text{F}(\mathbf{x}_m^{obs}) \cdot \text{G}(\mathbf{e}, \mathbf{d}))}{\sum_{(\tilde{\mathbf{e}}, \tilde{\mathbf{d}}) \in \mathcal{C}_\mathcal{F}} \exp(\text{F}(\mathbf{x}_m^{obs}) \cdot \text{G}(\tilde{\mathbf{e}}, \tilde{\mathbf{d}}))} ,\end{aligned} \quad (2)$$

where $\cdot$ denotes the dot-product operation, $\mathbf{e} \in \mathcal{E}_\mathcal{F}$ and $x_m^{obs} = x \setminus e_m^{corrupt}$.

For example, when the corrupted entity $\mathbf{e}_m^{corrupt}$ is "而在桓," and the original ASR hypothesis $\mathbf{x}$ is: "韓國媒體報導稱而在桓確實人在日本." After masking out the corrupted entity, $\mathbf{x}$ turns into $\mathbf{x}_m^{obs}$: "韓國媒體報導稱[MASK][MASK][MASK]確實人在日本."

### 2.4.1 Context Encoder

The context encoder, denoted by $\text{F}(\cdot)$, consists of $U$ layers of Transformer blocks and is followed by a span encoder layer. Before passing $\mathbf{x}_m^{obs}$ into the context encoder, we prepend two special tokens [ES] and [EE] to mark the beginning and end positions of the corrupted entity $\mathbf{e}_m^{corrupt}$ within $\mathbf{x}_m^{obs}$, which helps in extracting the context embedding of the masked entity $\mathbf{e}_m^{corrupt}$. For example, given that the original $\mathbf{x}_m^{obs}$ is "韓國媒體報導稱[MASK][MASK][MASK]確實人在日本." After inserting the special tokens, $\mathbf{x}_m^{obs}$ turns into "韓國媒體報導稱[ES][MASK][MASK][MASK][EE]確實人在日本." To extract the contextual embedding of the masked $\mathbf{e}_m^{corrupt}$, we first pass $\mathbf{x}_m^{obs}$ through the Transformer blocks, resulting in $\text{H} \in \mathbb{R}^{T \times d}$ which represents the $d$-dimensional embeddings of $T$ tokens in $\mathbf{x}_m^{obs}$. Let $\mathbf{h}_i \in \mathbb{R}^d$ be the hidden embedding of the $i$-th token. We then concatenate the embeddings of the [ES] and [EE] tokens and feed the resulting embedding into the span encoder to derive the final contextual embedding. The span encoder can be defined as follows:

$$\text{SpanEncoder}([\mathbf{h}_s; \mathbf{h}_e]) = \mathbf{W}[\mathbf{h}_s; \mathbf{h}_e], \quad (3)$$

where $s$ and $e$ denote the positions of the [ES] and [EE] tokens, respectively. $\mathbf{W}$ represents the weight matrix of a trainable linear transform, and $[\cdot\,;\,\cdot]$ serves as the concatenation of two embeddings.

---

[2] Pinyin is the romanization system for standard Mandarin Chinese.

| Model | AISHELL Test Set (%) | | | | Homophone Test Set (%) | | | |
|---|---|---|---|---|---|---|---|---|
| | CER | NNE CER | NE CER | NE Recall | CER | NNE CER | NE CER | NE Recall |
| Conformer | 4.62 | **4.00** | 11.12 | 78.36 | 8.41 | **5.27** | 15.58 | 70.25 |
| PED-NEC | 4.34 | 4.01 | 8.14 | 84.63 | 10.08 | 5.35 | 20.88 | 56.72 |
| - w/o rejection | 4.90 | 4.65 | 8.22 | 85.61 | 10.67 | 6.05 | 21.42 | 56.14 |
| DANCER | **4.29** | 4.00 | **7.57** | 85.85 | **7.17** | 5.30 | **11.33** | **79.84** |
| - w/o rejection | 4.84 | 4.64 | 7.63 | **86.81** | 7.87 | 6.12 | 12.04 | 78.68 |

Table 1: Main results of our DANCER model on the AISHELL-1 and Homophone test set.

### 2.4.2 Entity Encoder

The entity encoder $\mathrm{G}(\cdot)$ is constructed with $U$ layers of Transformer blocks. To obtain the semantic embedding of an entity, we concatenate its corresponding character tokens $\mathbf{e}$ and description tokens $\mathbf{d}$ together. Two special tokens, [CLS] and [SEP], are used to separate them. This forms the input sequence for the entity encoder. For example, when the entity $\mathbf{e}$ is "朴泰桓", and the description $\mathbf{d}$ is "韓國遊泳運動員出身於韓...", the input of the entity encoder turns into, " [CLS]朴泰桓[SEP]韓國遊泳運動員出身於韓...[SEP]." After passing the input sequence into $\mathrm{G}(\cdot)$, we extract the embedding of the [CLS] token as the final semantic representation $\mathrm{G}(\mathbf{e}, \mathbf{d})$ of the entity.

### 2.4.3 Training

The objective of EDA-MLM is to explicitly capture shared information between the context of the masked corrupted entity and the semantics of the correct entity description. Therefore, an effective approach is to maximize the mutual information between the contextual information residing in the contextualized representation of the masked corrupted entity $\mathrm{F}(\mathbf{x}_m^{obs})$ and the representation of the correct entity with its description $\mathrm{G}(\mathbf{e}, \mathbf{d})$.

| Model | Test Set NE-Recall (%) | | |
|---|---|---|---|
| | ≤ 0-shot | ≤ 5-shot | ≤ 100-shot |
| Conformer | 38.83 | 50.32 | 70.55 |
| PED-NEC | 61.73 | 66.92 | 79.82 |
| DANCER | **61.77** | **68.39** | **80.86** |

Table 2: Analysis of few-shot generalization ability on the AISHELL-1 test set.

Given a mini-batch of $B$ triples, denoted as $\mathcal{B} = \{(\mathbf{x}_i^{obs}, \mathbf{e}_i, \mathbf{d}_i)\}_{i \in \mathcal{I}}$, where $\mathcal{I} = \{1, \cdots, B\}$ represents the index of an arbitrary sample, and $\mathbf{x}_i^{obs}$ represents the masked ASR hypothesis along with its corresponding correct entity $\mathbf{e}_i$ and description $\mathbf{d}_i$. The mutual information between these two views can be denoted as $I(\mathrm{F}(\mathbf{x}_i^{obs}), \mathrm{G}(\mathbf{e}_i, \mathbf{d}_i))$. To maximize the mutual information, we employ the infoNCE loss (Oord et al., 2018). The infoNCE loss serves as an estimator of mutual information and can be represented as $I(\mathrm{F}(\mathbf{x}_i^{obs}), \mathrm{G}(\mathbf{e}_i, \mathbf{d}_i)) \leq \log(B) - \mathcal{L}_B$. Here, $B$ represents the size of the training samples, and minimizing the objective $\mathcal{L}_B$ is equivalent to maximizing the lower bound on the mutual information. The infoNEC loss can be expressed by:

$$\mathcal{L}_B = -\frac{1}{B} \sum_{i \in \mathcal{I}} \log \frac{\exp(\mathrm{F}(\mathbf{x}_i^{obs}) \cdot \mathrm{G}(\mathbf{e}_i, \mathbf{d}_i))}{\sum_{j \in \mathcal{I}} \exp(\mathrm{F}(\mathbf{x}_i^{obs}) \cdot \mathrm{G}(\mathbf{e}_j, \mathbf{d}_j))}. \quad (4)$$

### 2.5 Interpolation

We introduce a hyperparameter $\alpha$ to regulate the contributions of the phonetic and semantic scores more precisely, and we select the entity with the highest holistic score to substitute the corrupted entity. The holistic score can be computed by: $\alpha \log P_\eta(\mathbf{e} \mid \mathbf{e}_m^{corrupt}) + (1-\alpha) \log P_\theta(\mathbf{e} \mid \mathbf{x}_m^{obs})$.

### 2.6 Entity Rejection

We incorporate an entity rejection mechanism into our DANCER model to ensure that it can avoid mistakenly replacing named entities that are already correctly recognized. Our entity rejection operates in a manner similar to (Garg et al., 2020) and utilizes the *n*-best ASR hypotheses to determine whether to accept or reject an entity candidate generated by the DANCER model. The text span of the entity in each ASR hypothesis provides additional phonetic information about the original audio. If the phonetic information of the entity candidate proposed by the DANCER model significantly differs from that of the entity span in the corresponding locations of all *n*-best hypotheses, it should be rejected since it might deviate from the original audio.

Given the corrupted entities $\mathcal{E}_1^{corrupt}$ in the top-one hypothesis detected by the corrupted entity detector (CED), we first align the corresponding corrupted entities with the other top-*n* hypotheses. For example, $\mathcal{E}_n^{corrupt} = (\mathbf{e}_{1,n}^{corrupt}, \cdots, \mathbf{e}_{M,n}^{corrupt})$ represents the corrupted entities in the *n*-th best ASR hypothesis, with $\mathbf{e}_{m,n}^{corrupt}$ denoting the *m*-th corrupted entity within it. The rejection score can be calculated as follows:

$$\text{Reject}(\mathbf{e}_m) = \sum_{n \in \mathcal{N}} p_n \cdot \text{ED}(\text{Pin}(\mathbf{e}_{m,n}^{corrupt}), \text{Pin}(\mathbf{e}_m)), \quad (6)$$

where $p_n$ represents the beam search score of the top *n*-th ASR hypothesis, and $\mathcal{N} = \{1, \cdots, N\}$ signifies the index of any arbitrary hypothesis in the *n*-best list. If the rejection score $\text{Reject}(\hat{\mathbf{e}}_m)$ for the entity hypothesis $\hat{\mathbf{e}}_m$ generated by the DANCER model is higher than the rejection score $\text{Reject}(\mathbf{e}_{m,1}^{corrupt})$ of the original corrupted entity $\mathbf{e}_{m,1}^{corrupt}$ proposed by the CED model, we reject replacing the corrupted entity with $\hat{\mathbf{e}}_m$.

## 3. Experiments Setup

### 3.1 Dataset and Evaluation Metrics

#### 3.1.1 Dataset

Our experiments were conducted on AISHELL-1 and a Homophone test set containing high phonetic confusion data sampled from AISHELL-1. AISHELL-1 is a commonly used open-source speech corpus for evaluating Chinese ASR systems, containing over 170 hours of Mandarin speech data across diverse domains, such as "Finance," "Science and Technology," "Sports," "Entertainment," and "News." We also utilized the AISHELL-NER dataset (Chen et al., 2022), which was constructed based on AISHELL-1, to obtain the tagging information of named entities. The NE list $\mathcal{E}$ was initially constructed from the whole training, development, and test sets of AISHELL-1. After filtering out the entities where we could not find their corresponding descriptions (*cf.* section

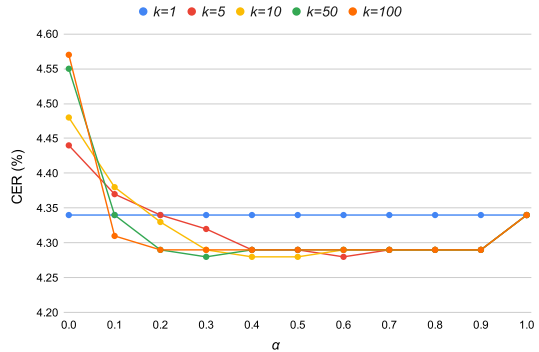

Figure 4: Impact of the different settings of $\alpha$ and top-*k* on the AISHELL-1 test set.

3.2), the NE list $\mathcal{E}$ contained 16,168 distinct named entities.

To construct the Homophone test set, we followed two procedures using the AISHELL-1 dataset to sample high phonetic confused utterances. Firstly, we applied the Pinyin transformation and the normalized edit distance algorithm to calculate the pair-wise phonetic similarity score for any two entities in the NE list $\mathcal{E}$. If the similarity score between two entities was equal to 1, we identified them as a pair of phonetically confusing entities. In the second step, we collected all distinct phonetic confusing entities from the NE list. Next, we curated the utterances containing these confusing entities from the AISHELL-1 test set to form the Homophone test set. The final Homophone test set contains 115 highly phonetically confusing speech utterances.

#### 3.1.2 Evaluation Metrics

We will evaluate the performance levels of disparate NEC methods with respect to four metrics: 1) CER: overall character error rate on the whole test set; 2) NNE-CER: character error rate for the non-entity characters in the utterance; 3) NE-CER: character error rate for entities characters in the utterance; and 4) NE-Recall: the recall of the entity in the utterance that are correctly recognized.

### 3.2 Entity Description Construction

To construct entity-description pairs $\mathcal{C}$, we utilized Chinese Wikipedia and Baidu Baike as our sources of knowledge. Initially, we used a given entity as the query to search for the most relevant article from these two sources. Subsequently, we applied a text normalization process to eliminate semi-structured data from the acquired article, such as those composed of only information boxes, tables, and lists. The entity description was then formed by extracting the first 100 words from the article, as the initial few words generally serve as the abstract and can meet our requirements.

## 3.3 Baseline and Model Configuration

### 3.3.1 Baseline

Our ASR model comprised a Conformer encoder (Gulati et al., 2020) and a Transformer decoder (Vaswani et al., 2017) (denoted by Conformer for short), trained on the training set of AISHELL-1 using the ESPnet toolkit[3]. We utilized the same hyperparameter setting as the recipe provided in the toolkit. The method proposed in this paper will be compared against the phonetic edit-distance-based named entity corrector (PED-NEC) method (Raghuvanshi et al., 2019) (with phonetic features and *n*-best list), which is taken as the strong baseline in this paper.

### 3.3.2 Model Configuration

We trained each component in the DANCER model separately and utilized the *n*-best ASR transcripts of the utterances in the training set of AISHELL-1 as additional noisy data to enhance our system. Here, we set $N = 10$. In the entity description augmented MLM model (EDA-MLM), we used 12-layer Transformer blocks initialized with the BERT base pre-trained parameters for use in the context encoder and the entity encoder. We employed a single linear transform with 768 hidden dimensions for the span encoder layer.

To construct the training set for EDA-MLM, we utilized the *n*-best transcripts of the utterances in the training set of AISHELL-1 to form the masked utterance $\mathbf{x}_i^{obs}$ and the corresponding correct entity $\mathbf{e}_i^{corrupt}$. We used open-sourced DPR toolkit[4] for training our EDA-MLM model with a batch size $B = 32$. Before evaluation, we extracted all the entity-description embeddings from the $\mathcal{C}$ using the entity encoder $\mathrm{G}(\cdot)$ and formed the entity-description memories. An efficient inner product search method was employed to accelerate the masked prediction procedure during inference.

For the corrupted entity detector (CED), we utilized the same BERT base architecture. As there were no corrupted entity datasets available, we first assigned the BIO tag to the n-best hypotheses from AISHELL-1 training set by aligning them with the manuscript. Then, we utilize the aligned hypotheses with their corresponding BIO tag to form our corrupted entity dataset. When tested on the hypotheses of AISHELL-1 test-set transcribed by the ASR model, the Precision, Recall and F1 score of the CED model is 91.49%, 92.66% and 91.52%, respectively. Finally, through an empirical study on the effect of alpha $\alpha$ and top-*k* using the development set, we set $\alpha = 0.6$ and $k = 10$. For the baseline model PED-NEC, we utilize the same CED model, phonetic-level retrieval, and entity rejector settings as in DANCER.

## 4. Experimental Results

### 4.1 Main Results

The main results shown in Table 1 demonstrate that our approach, which utilizes the semantics of the entity, leads to a better reduction in CER for both datasets. Particularly on the Homophone test set, which contains highly phonetic-confused data, our DANCER model achieved an additional character error rate (CER) reduction of 28.87% relatively over the phonetic edit-distance-based NEC (PED-NEC) model.

Incorporating the entity rejection mechanism may slightly decrease the NE recall rate. The inclusion of this cautious process allows our model to significantly reduce the CER pertaining to the non-entity parts of test utterances. As such, it benefits the overall CER, which could otherwise be adversely affected by errors arising from the misidentification made by the corrupted entity detector.

### 4.2 Detailed Analysis

#### 4.2.1 Few-shot Generalization

Table 2 displays the few-shot generalization ability of both the baseline and our proposed models. In this context, 0-shot refers to unseen entities during model training, while 5-shot and 100-shot indicate entities that appeared within five and one hundred times, respectively, in the utterances of the training set. Our DANCER model consistently outperforms the PED-NEC model. Additionally, our EDA-MLM module can effectively adapt and accommodate to unseen entities by leveraging the entity-description memories, demonstrating promising zero-shot ability.

#### 4.2.2 Impact of the Entity List Size

This subsection further showcases that when the NE list grows, the problem incurred by phonetic confusion in the NE list will increase substantially. When scaling up the NE list, as shown in Figure 1, we observe that PED-NEC decreases the entity recall rate on the AISHELL-1 test set more severely than DANCER. The gap between PED-NEC and DANCER increases as the size of the entity list increases.

#### 4.2.3 Impact of the Different Settings of the $\alpha$ and Top-*k*

Figure 4 illustrates an empirical study conducted on the AISHELL-1 test set to examine the impact of different combinations of $\alpha$ and top-*k* values on the overall CER performance. The hyperparameter $\alpha$ controls the percentage of the phonetic similarity score with the semantic similarity score predicted by the EDA-MLM. Meanwhile, top-*k* deter-mines the initial size of the entity candidates selected by the phonetic-level retriever. As can be observed from Figure 4, incorporating the semantic information of entities, wherein $(1-\alpha)$ is greater than zero, reduces the overall CER. The combination of equally-weighted phonetic and semantic scores appears to best balance their effects. However, increasing the number of retrieved entities (namely, value of *k*) leads to increasing instability, which could be attributed to the inclusion of entities that deviate significantly from the input utterance. Therefore, setting the value *k* equal to 10 seems to be a reasonable choice.

---

[3] https://github.com/espnet/espnet

[4] https://github.com/facebookresearch/DPR

## 5. Conclusion and Future Work

In this paper, we have designed and implemented a novel entity description augmented named entity corrector for ASR, dubbed DANCER, which leverages entity descriptions to provide additional information that helps mitigate the problems of phonetic confusion incurred by ASR NEC. Specifically, an efficient entity description augmented masked language model (EDA-MLM) has been proposed to incorporate a dense retriever and entity-description memories to enable adapting and accommodating ASR NEC to domain-specific entities. Empirical experiments conducted on the AISHELL-1 and Homophone datasets validate the effectiveness of our approach. In future work, we plan to explore alternative entity modeling regimes, such as graph-based modeling, and incorporate the NE list ahead of time into the corrupted entity detector to reduce the search space since we care more about the corrupted entities whose correct counterparts are included in the NE list for practical use cases.

## Limitations

While our model achieves better entity correction performance over PED-NEC, it still has several limitations. Firstly, the entity description needs to be defined in advance to help alleviate the phonetic confusion in NEC model. Second, the correction ability of DANCER is bounded by the detection quality of the CED model. Though effective, more sophisticated corrupted entity detection methods are expected to further improve the performance, which can be a future direction.

## Acknowledgements

This work was supported in part by E.SUN Bank under Grant Number 202308-NTU-02. Any findings and implications in the paper do not necessarily reflect those of the sponsors.

## 6. Bibliographical References